 \documentclass[final,3p,times,twocolumn]{elsarticle}

\usepackage[utf8]{inputenc}
\usepackage[T1]{fontenc}
\usepackage{graphicx}
\usepackage{amssymb}
\usepackage{amsthm}
\usepackage[fleqn]{amsmath}
\usepackage{lineno}
\usepackage{caption}
\usepackage{txfonts}
\usepackage{array}
\usepackage{booktabs}
\usepackage{setspace}
\usepackage{hyperref}
\usepackage{url}

\usepackage[font=small,labelfont=bf,labelsep=none]{caption}
\begin{document}

\begin{frontmatter}


\title{Point Clouds Learning with Attention-based Graph Convolution Networks}


\author[fn1]{Zhuyang Xie}
\ead{zyxie@my.swjtu.edu.cn}
\author[fn2]{Junzhou Chen\corref{cor1}}
\ead{chenjunzhou@mail.sysu.edu.cn}
\author[fn1]{Bo Peng}
\ead{bpeng@swjtu.edu.cn}

\address[fn1]{School of Information Science and Technology, Southwest Jiaotong University, Chengdu, Sichuan 611756, China}
\address[fn2]{Research Center of Intelligent Transportation System, School of Intelligent Systems Engineering, Sun Yat-sen University, Guangzhou, Guangdong 510006, China}
\cortext[cor1]{Corresponding author.}

\begin{abstract}
Point clouds data, as one kind of representation of 3D objects, are the most primitive output obtained by 3D sensors. Unlike 2D images, point clouds are disordered and unstructured. Hence it is not straightforward to apply classification techniques such as the convolution neural network to point clouds analysis directly. To solve this problem, we propose a novel network structure, named Attention-based Graph Convolution Networks (AGCN), to extract point clouds features. Taking the learning process as a message propagation between adjacent points, we introduce an attention mechanism to AGCN for analyzing the relationships between local features of the points. In addition, we introduce an additional global graph structure network to compensate for the relative information of the individual points in the graph structure network. The proposed network is also extended to an encoder-decoder structure for segmentation tasks. Experimental results show that the proposed network can achieve state-of-the-art performance in both classification and segmentation tasks.
\end{abstract}


\end{frontmatter}


\section{Introduction}
With the rapid development of data acquirement techniques, 3D sensors have been widely used in robotics, autonomous, reverse engineering and virtual reality. There are increasing demands for 3D data analysis algorithms \cite{Charles2017PointNet,Qi2017PointNet,wang2018local,xu2018spidercnn,xie2018attentional,wang2015voting}, meanwhile a large number of 3D point clouds datasets are available recently ~\cite{Wu20143D,armeni2017joint,chang2015shapenet,Dai2017ScanNet,Hackel2017Semantic3D}. Due to the irregular distribution in 3D space and the lacking of canonical order, 3D point clouds data are difficult to be processed by traditional methods.\\
\indent In  recent years, convolutional neural networks (CNNs) have achieved a great success in processing the standard grid data for tasks such as image recognition~\cite{he2016deep,ioffe2015batch}, semantic segmentation~\cite{long2015fully,Noh2015Learning} and machine translation~\cite{cheng2016long,vaswani2017attention}. By virtue of the success of CNNs, many methods focus on 3D voxels~\cite{Maturana2015VoxNet,Riegler2017OctNet,Qi2016Volumetric}, which convert point clouds into 3D volumetric grid in the pre-processing step. In this way, features from voxels can be extracted by the 3D convolution network. However, the calculations may be redundant due to the sparsity of most 3D data and its computational complexity increases exponentially with the increase of resolution~\cite{Maturana2015VoxNet,Riegler2017OctNet,Qi2016Volumetric,li2016fpnn}.\\
\indent Some latest work has focused on direct learning on point clouds. As a pioneer of directly processing point clouds, PointNet~\cite{Charles2017PointNet} provides an effective strategy to learn the features of individual points through a shared multi-layer perception (MLP), and eventually encodes global information by a symmetric function that guarantees permutation invariance to the points’ order. However, the MLP of PointNet~\cite{Charles2017PointNet} is based on the feature learning of individual points, and does not consider the local geometry. To solve this problem, PointNet++~\cite{Qi2017PointNet} divides the point set into several subsets, sends these subsets to a shared PointNet, and builds a hierarchical network by repeating such a process iteratively. Although PointNet++~\cite{Qi2017PointNet} has built local point sets, the relationship between these local point sets is not well constructed. In the latest work, ShapeContextNet~\cite{xie2018attentional} uses the self-attention mechanism to learn the relationship between individual points. However, they regards attention as a query operation, and calculates attention score for each individual point on the whole point clouds, which significantly increases the computation cost and memory usage.\\
\indent In order to solve the above problems, we mainly consider two aspects, that is, how to model the relationship between local structural information and how to effectively aggregate local information. In this paper, we propose AGCN to learn point clouds feature. Specifically, in the stage of local structure learning, some nodes are first obtained by sparse sampling and for each node we construct a local point set. Then the local structure feature of each node is directly learned on its local point set. Subsequently, we construct a KNN graph for these nodes, and design a point attention layer on the KNN graph to learn the relationship between different local information and gather the information of neighbors for each node. In our network, we can learn from the local to the global features by stacking multiple layers of point attention layer. Also AGCN can propagate high-level features to the fine-grained features and we extended our network to an encoder-decoder structure for segmentation tasks. In addition, in order to compensate for the relative information of the individual nodes in the graph structure network, we propose a global point graph to assist the learning of point attention layer. Our key contributions are as follows:
\begin{itemize}
  \item We propose a point attention layer on the KNN graph to calculate the attention score of the $K$ nearest neighbors, which can effectively aggregate the local information and guarantee permutation invariance to the points' order.
  \item We propose a global point graph to compensate for the relative location information of the individual nodes in the graph structure network.
  \item We extend point attention layer and propose an attention-based encoder-decoder network for point clouds segmentation.
  \item We have achieved better performance on some standard datasets compared with state-of-the-art approaches.
\end{itemize}

\section{Related Works}
In this section, we will briefly review some of the current approaches for 3D data. Specifically, they can be classified into multi-view based methods, voxel based methods, graph based methods, and point based methods.

\subsection{View-based Methods}
The view-based approaches~\cite{su2015multi,Novotny2017Learning} project a 3D object into a collection of 2D views, which applies the conventional 2D convolutional neural network to each view, and then aggregate these features by multi-view pooling for classification and retrieval tasks. However, the view-based method requires a complete view set for each target, which adds preprocessing work and computation cost. Also, it is nontrivial to extend them to scene understanding or other 3D tasks(e.g., per-point classification), because the view-based approaches lose 3D spatial information.

\subsection{Volumetric Methods}
The voxelization methods converts unstructured geometric data into 3D regular grid~\cite{Maturana2015VoxNet,Riegler2017OctNet,Qi2016Volumetric}, which can be applied to 3D convolution operation. However, the volume representation is often redundant due to the sparsity of most 3D data. Therefore, a voxel method is usually limited by the resolution of volumetric grids and the computation cost of 3D convolution, which leads to use lower resolution as input and it is hard to learn local geometric details. In addition, due to the limitation of resolution, it is challenging for a voxel method to process large scale point clouds data.

\subsection{Graph-based Methods}
The use of graphs to represent irregular or non-European data (such as point clouds, social networks) are flexible. There are two classes of methods of this kind. The first method directly defines convolution operation on the graph. ECC~\cite{Simonovsky2017Dynamic} is the first work to apply the graph convolution to point clouds. It defines filter weights conditioned on the specific edge labels in the neighborhood of a vertex. KC-Net~\cite{shen2018mining} contains a KNN graph to extract the local structural feature of the point clouds and aggregates the neighbor information through the graph max pooling. DGCNN~\cite{wang2018dynamic} proposes EdgeConv on a KNN graph to achieve local information fusion by learning the features of the edges of neighbor points.\\
\indent The other is spectral based methods, which define the convolution operations in the Fourier domain~\cite{yi2017syncspeccnn,Kipf2016Semi,Defferrard2016Convolutional}. Some latest work has shown concern in this area. SpiderCNN~\cite{xu2018spidercnn} defines a series of convolution kernels, as a product of a simple step function and a Taylor polynomial, to approximate the weight function. LocalSpecGCN~\cite{wang2018local} use spectral convolution combined with recursive clustering and pooling strategy to extract features of neighbor points. However, spectral based methods cause a large number of parameters of the convolution filter.

\subsection{PointNets}
\begin{figure*}[ht]
    \centering
    \begin{minipage}[c]{0.95\textwidth}
    \includegraphics[width=\textwidth]{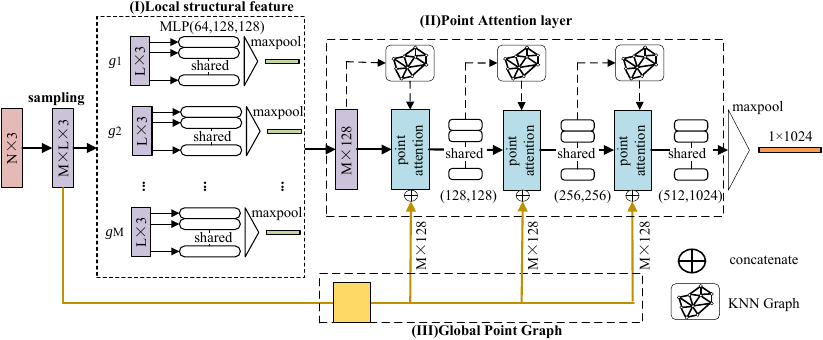}
    \end{minipage}\hfill
    \begin{minipage}[c]{\textwidth}
	\caption{The architecture of AGCN. AGCN firstly samples $M$ nodes based on the input point clouds, then extracts local point sets of size $L$ for each node, and learns local features for each node (I). The KNN graph is constructed according to the coordinates of $M$ nodes, and the feature aggregation of neighbor nodes is realized by introducing the attention mechanism in the KNN graph (II), we stacked 3 layers of point attention layer for classification. In order to compensate for the relative information of the individual nodes in the graph structure network, we additionally construct a global graph structure network to assist the learning of the point attention layer (III).}
    \label{Figure1:framework}
    \end{minipage}
\end{figure*}

Some recent work has focused on learning directly from point clouds. PointNet~\cite{Charles2017PointNet} uses MLP with shared weights to learn the features of individual points. Finally, the point clouds information is coded by maxpool for classification. However, PointNet~\cite{Charles2017PointNet} uses the individual points' feature, and does not utilize local information. To solve this problem, PointNet++~\cite{Qi2017PointNet} builds a hierarchical network and extracts multi-scale information from local point sets. The different hierarchical information is learned by iteratively radius search, but this also increases the computational complexity and does not model the relationship between different local information. Kd-Net~\cite{Klokov2017Escape} constructs a hierarchical structure through kd tree, divides space according to three axes and learns the weights along specific axis. However, as the number of point clouds and the resolution of kd tree increase, the corresponding computation cost will also increase.\\
\indent There is also a part of work to learn the local features of the point clouds by constructing convolution kernels. KC-Net~\cite{shen2018mining} defines a series of learnable point sets on a KNN graph by kernel correlation, which is used to extract the local structure feature of point clouds. ShapeContextNet~\cite{xie2018attentional} constructs shape context kernels, through the concept of shape context. In addition, in order to deal with point clouds which usually have varying size and density, the A-SCN~\cite{xie2018attentional} network based on self-attention is proposed.\\
\indent In our work, we mainly focus on and learn the relationship between different local information. Inspired by KC-Net~\cite{shen2018mining}, we build KNN graph to learn each nodes' local structure feature, and implement local feature aggregation through the attention mechanism, which is different from graph max pooling used in KC-Net~\cite{shen2018mining}. Unlike A-SCN~\cite{xie2018attentional}, in point attention layer, we only calculate the attention score for $K$ neighbors, which greatly reduces the computation cost. And our proposed point attention layer can be stacked in multiple layers for better classification and segmentation tasks.

\section{Method}
Our method mainly consists of three parts: (1)learning local structural feature (Section~\ref{sec3.1:local_structural_learning}); (2)point attention layer (Section~\ref{sec3.2:point_attention_layer}); (3)global point graph (Section~\ref{sec3.3:global_graph}). Figure~\ref{Figure1:framework} illustrates our full network architectures for classification.

\subsection{Learning Local Structural feature}
\label{sec3.1:local_structural_learning}
The input of the original point clouds are represented by three-dimensional coordinates $\mathbf{P}=\{p_{i}\in\mathbb{R}^{3},i=1,2,...,N \}$, and the features such as color, surface normal or other information can also be added. We extract local point sets in the same way as PointNet++~\cite{Qi2017PointNet} does. In Figure~\ref{Figure1:framework}(I), $M$  nodes are sampled from the farthest point in $\mathbf{P}$ , which forms a set $\mathbf{S}=\{ s_j \in \mathbb{R}^{3},i=1,2,...,M \}$, and $\mathbf{S}\subseteq\mathbf{P}$. For each node $s_j$, a local point set $\mathbf{G}=\{ g_j \in \mathbb{R}^{L\times3},j=1,2,…,M \}$ is constructed, where $g_j$ is a local point set obtained by taking the node $s_j$ as the center, and extracting $L$ points of the nearest neighbor of the node $s_j$. We extract the local structural features for each point set $g_j$ by learning a local mapping $\emph{f}:\mathbb{R}^{L\times3}\rightarrow\mathbb{R}^{\emph{d}}$, which converts local point set to a \emph{d}-dimensional vector. The specific function $\emph{f}$ is defined as:
\begin{equation}
\centering
\emph{f}(g_j)=\mathit{\rm{max}}_{l=1,2,...,\rm{L}}(\mathit{\rm{MLP}}(g_{j,l}-s_j)),\label{equ:1}
\end{equation}
where $g_{j,l}$ represents the $l$-th point of the local point set $g_j$, $g_{j,l}-s_j$ represents the normalized coordinate, and the features of the individual points are extracted by three MLPs with shared weights. Finally, the features of individual points are fused by local maxpool.\\
\indent By constructing the local point set $\mathbf{G}$, we learn the local feature representation for each node $s_j$, and use the features of $M$ nodes as the input of the subsequent network for feature learning, which further reduces the computation cost of the latter network.

\subsection{Point Attention layer}
\label{sec3.2:point_attention_layer}
Attention mechanism is widely used in different types of deep learning tasks such as natural language processing~\cite{vaswani2017attention,cheng2016long}, modeling the relationship about relevant parts. In this section, we will introduce the attention-based point attention layer, to learn the relationship between adjacent points. By learning the local structural features of Section~\ref{sec3.1:local_structural_learning}, we obtain the local feature representation of $M$ nodes: $\mathbf{F}=\{ f_m|m=1,2,…,M \}, \emph{f}_{\emph{m}}\in{\mathbb{R}}^{1\times{\emph{d}}}$. The features of these nodes are taken as the input of point attention layer, and the updated features of these nodes are obtained as the output of the network: $\mathbf{F}'=\{ f_m'|m=1,2,…,M \},\emph{f}_{\emph{m}}'\in\mathbb{R}^{1\times{\emph{d}}'}$.

\begin{figure}[!ht]
  \begin{minipage}[c]{0.48\textwidth}
  \includegraphics[width=\textwidth]{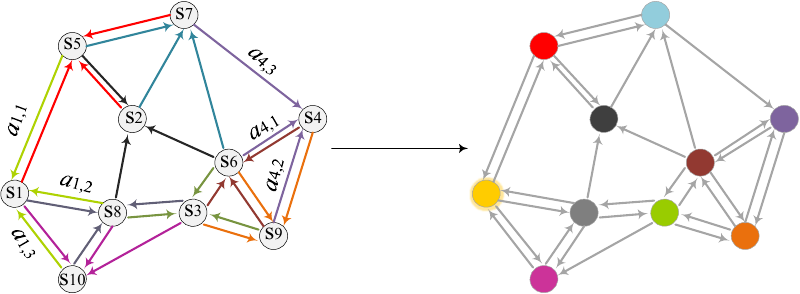}
  \end{minipage}\hfill
  \begin{minipage}[c]{0.48\textwidth}
	\caption{Point attention layer. We illustrate a point attention layer with 3-NN graph. In the left part of this figure, for each node $s_j$, we aggregate the information of the neighbors around the node $s_j$ according to the attention score. Arrows indicate the direction of the information propergation, and different colors indicate independent attention calculations. In the right part of this figure, each node with different color represents the aggregated feature.}
	\label{Figure2:attention_layer}
  \end{minipage}
\end{figure}

\indent As illustrated in Figure~\ref{Figure2:attention_layer}, we construct a KNN graph for $M$ nodes. Different from KC-Net's~\cite{shen2018mining} graph max pooling for neighbor information aggregation, we focus on $K$ nearest neighbor nodes around the node $s_j$ and aggregate the information of the neighbor nodes according to the attention score. The feature aggregation formula of node $s_j$ is as follows:
\begin{equation}
\centering
\emph{f}_{j}'=\sum_{\emph{k}\in\mathcal{N}(s_j)} \alpha_{j,k} \cdot \emph{f}_{j,k} + \emph{f}_j,\label{equ:2}
\end{equation}
where $f_j$ denotes the feature of node $s_j$. $\mathcal{N}(s_j)$ denotes the index of the neighbors of node $s_j$. $f_{j,k}$ denotes the feature of the $k$-th nearest neighbor of node $s_j$, and $f_{j}'$ indicates the updated feature of node $s_j$. $\alpha_{j,k}$ denotes the attention score between node $s_j$ and the $k$-th neighbor. The $\sum$ operation can be regarded as a weighted summation of the $K$ neighbor nodes around node $s_j$, which guarantees the permutation invariance to the nodes' order. Attention $\alpha_{j,k}$ is calculated as follows:
\begin{equation}
\centering
\alpha_{j,k}=\frac{\emph{f}_{j}^{\rm{T}} \cdot \emph{f}_{j,k}} {\sum_{\emph{k}\in\mathcal{N}(s_j)} \emph{f}_{j}^{\rm{T}} \cdot \emph{f}_{j,k}},\label{equ:3}
\end{equation}
\indent With Equation (~\ref{equ:2}), each individual node's feature can be updated in parallel. In addition, in order to incorporate additional nonlinearity and increase the capacity of the model, we add a feature transformation function, which uses a 2-layer MLP with a nonlinear activation function to perform feature transformation on each updated feature $f_{j}'$.\\
\indent As shown in Figure~\ref{Figure1:framework}(II), in our network structure, a multi-layered point attention layer is adopted, which is a very effective structure. By stacking multiple layers of point attention layer, a CNN-like effect can be achieved. The number $K$ of neighbor nodes can be regarded as the kernel size in CNN, and as the network depth increases, the receptive field of the network increases correspondingly, so that from local to global information can be learned.

\subsection{Global Point Graph}
\label{sec3.3:global_graph}
\begin{figure*}[t]
    \centering
    \begin{minipage}[c]{0.9\textwidth}
    \includegraphics[width=\textwidth]{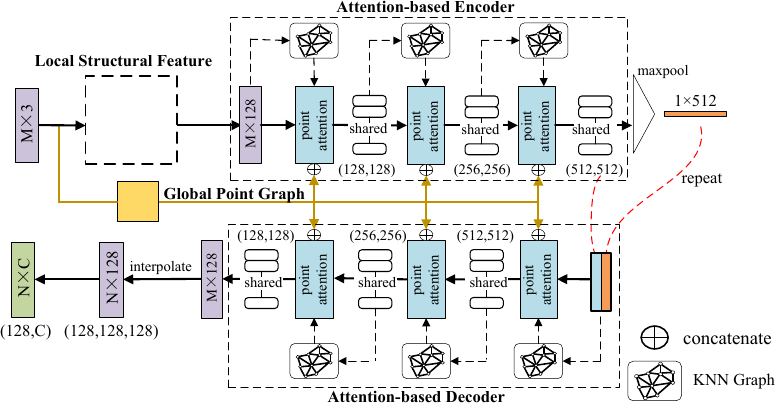}
    \end{minipage}\hfill
    \begin{minipage}[c]{\textwidth}
	\caption{Attention-based encoder-decoder architecture. The encoder-decoder architecture is an invert operation. The encoder aggregates neighbor information through attention, and the decoder propagates high-level semantic information to lower-level finer information. All nodes' features of global point graph are concated with corresponding nodes' features in each point attention layer.}
    \label{Figure4:Encoder-Decoder}
    \end{minipage}
\end{figure*}

\begin{figure}[!ht]
  \begin{minipage}[c]{0.49\textwidth}
  \includegraphics[width=\textwidth]{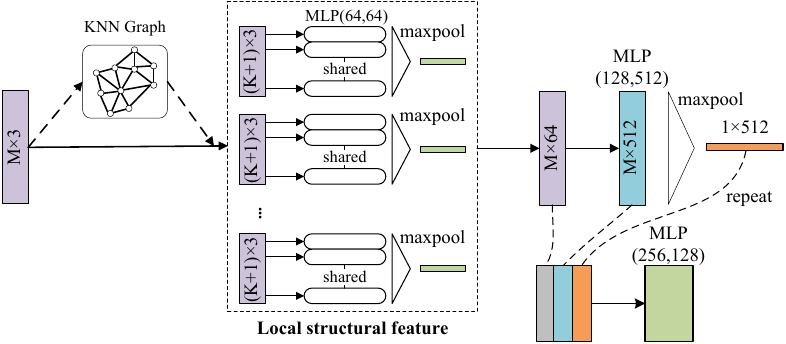}
  \end{minipage}\hfill
  \begin{minipage}[c]{0.49\textwidth}
	\caption{Global point graph. We construct a KNN graph for $M$ nodes. ($K$+1) denotes $K$ nearest neighbor nodes and node $s_j$. We get the global graph feature through maxpool, and concat the global feature with individual nodes to learn each node's relative information.}
	\label{Figure3:global_graph}
  \end{minipage}
\end{figure}

In our entire network, nodes' features can only describe local features and do not provide information relative to the global field. So we design a simple network to build a global structure diagram to learn the global information of each node. As shown in Figure~\ref{Figure3:global_graph}, the entire network can be seen as a simplified PointNet~\cite{Charles2017PointNet}, taking the sampled $M$ nodes as input. The same as the Section~\ref{sec3.2:point_attention_layer}, we construct a KNN graph and the local feature of each node $s_j$ is learned by a two-layer MLP. Finally, we get the global feature through maxpool. We concat the global feature with features of the individual nodes to learn the representation of each node relative to the global.\\
\indent In Figure~\ref{Figure1:framework}(III), we concat each node's feature learned by the global point graph with the corresponding nodes' features $f_{j}'$ in the point attention layer, giving these nodes global information. The experimental result shows that the global point graph can further combine the local information with the global information, thus assisting the learning of point attention layer and improving the performance of the network.

\subsection{Attention-based Encoder-Decoder for Segmentation}
\label{sec3.4:encoder-decoder}

In the classification network, from the local structure feature learning to the multi-layer point attention layer, and the global maxpool, the whole process can be regarded as an encoder. In the task of point-by-point classification, such as segmentation, requires the integration of local and global information, and the integration process can be regarded as an invert operation of the encoder. Therefore, we design a decoder network which illustrated in Figure~\ref{Figure4:Encoder-Decoder}. In the decoder, we use the attention structure opposite to the encoder to concatenate the local information of $M$ nodes with the global feature as the input of the decoder network.\\
\indent Intuitively, the encoder aggregates neighbor information through attention to achieve feature learning from local to global. On the contrary, in the decoder, it can be seen that each node sends the global information to its' neighbor nodes, which makes high-level semantic information propagate to the finer information.\\
\indent Finally, in order to obtain more fine-grained local information for the segmentation, we use the inverse distance weighted average interpolation in 3D Euclidean space. The formulas are as follows:
\begin{equation}
\centering
\emph{f}_i=\frac{\sum_{j=1}^{m} \emph{w}_{i,j} \cdot \emph{f}_{i,j} }{\sum_{j=1}^{m} \emph{w}_{i,j}},\label{equ:4}
\end{equation}
where \emph{m} represents \emph{m} nearest neighbor points in 3D Euclidean space, and \emph{m}=3 in the experimental setup; $\emph{w}_{(i,j)}=\frac{1}{\emph{d}(\emph{p}_i-\emph{p}_j)}$ represents the inverse square Euclidean distance between point $p_i$ and the neighbor point $p_j$.

\section{Experimental results}
To verify the performance our AGCN network, we compare some point-based methods on classification and segmentation tasks. The data sets
used mainly include ModelNet40~\cite{Wu20143D}, ShapeNet part dataset~\cite{chang2015shapenet}, and Large-Scale 3D Indoor Spaces Dataset (S3DIS)~\cite{Armeni20163D}.

\subsection{3D Point Set Classification}
\label{sec4.1:3D_cls}
We evaluate our network on ModelNet40~\cite{Wu20143D} for 3D point set classification. ModelNet40 contains 12311 CAD models from 40
categories and is split into 9843 for training and 2468 for testing. For fair comparison, we use the same data provided by PointNet
~\cite{Charles2017PointNet}. In our experiment, we employ the same augmentation strategy as PointNet~\cite{Charles2017PointNet} by
randomly rotating point clouds along the z-axis and jittering the position of each point by a Gaussian noise with zero mean and 0.02
standard deviation.\\
\indent As we described in Section~\ref{sec3.1:local_structural_learning}, our network uses the coordinates of the point clouds
as well as the surface normal as input. In the stage of local structural feature learning, we sample $M$=256 nodes to form
a set $\mathbf{S}$, and extract $L$=16 nearest points for each node $s_j$ to build local point set $g_j$. In the encoder, we
stack 3 layers of point attention layer and build a 3-NN graph for each point attention layer to learn from local to global
features. In addition, we build a 3-NN graph for the point set $\mathbf{S}$ as the input to the global point graph. Finally,
we get a global feature through maxpool and send it to three fully connected layers: $\rm{FC}(512)\rightarrow \rm{FC}(256) \rightarrow \rm{FC}(40)$ for object classification. Dropout layers are used for the fully connected layers, dropout ratio is 0.5. ReLU and Batchnorm are used in each MLP layer. In our network, all parameters are uniformly initialized within [-0.001, 0.001]. We
train the network for 200 epochs on an NVIDIA GTX 1080 GPU using tensorflow with Adam optimizer, batchsize=32, and initial learning
rate is 1e-3, momentum=0.9, the learning rate is reduced by a decay rate of 0.7 for every 20 epochs.\\
\indent In Table~\ref{Table1:ModelNet40_cls}, compard with the current methods, we can see that the proposed  method has achieved
state-of-the-art performance among the point-based methods.

\begin{table}[h]
\setlength{\abovecaptionskip}{0pt}
\setlength{\belowcaptionskip}{10pt}
\caption{Classification accuracy (\%) on ModelNet40.}\label{Table1:ModelNet40_cls}
\parbox{\textwidth}{
\begin{tabular}{l  c  c  c}
    \hline
    Method & input & accuracy & accuracy \\
     & points & avg.class & overall \\
    \hline
    ECC~\cite{Simonovsky2017Dynamic} & 1024$\times3$ &83.2 &87.4 \\
    PointNet~\cite{Charles2017PointNet} & 1024$\times3$ & 86.2 & 89.2 \\
    A-SCN~\cite{xie2018attentional} & 1024$\times3$ &87.4 &89.8 \\
    KC-Net~\cite{shen2018mining} & 1024$\times3$ & - & 91.0 \\
    PointNet++~\cite{Qi2017PointNet} & 5000$\times6$ & - &91.9 \\
    Kd-Net~\cite{Klokov2017Escape} & $2^{15}\times3$ & - & 91.8 \\
    SpiderCNN~\cite{xu2018spidercnn} & 1024$\times6$ & - & 92.4 \\
    LocalSpecGCN~\cite{wang2018local} & 2048$\times6$ & - & 92.1 \\
    \hline
    AGCN & 1024$\times6$ & \textbf{90.7} & \textbf{92.6} \\
    \hline
\end{tabular}}
\end{table}

\subsection{2D Point Set Classification}
\label{sec4.2:2D_cls}
We also evaluate the performance of the network on the MNIST dataset. We use the same protocol as used in PointNet++~\cite{Qi2017PointNet},
where 512 points are sampled for each digit image. Our network uses the coordinates of the 2D point clouds, we sample $M$=128 nodes
to form a set $\mathbf{S}$, and extract $L$=32 nearest points for each node $s_j$ to build local point set $g_j$. The other
experimental settings are the same as section~\ref{sec4.1:3D_cls}. Table~\ref{Table2:MNIST_cls} shows the classification result of our
network, and we can see that our network can achieve performance compared to the most recent methods on the 2D dataset.

\begin{table}[ht]
    \setlength{\abovecaptionskip}{0pt}
    \setlength{\belowcaptionskip}{10pt}
    \caption{Error rate (\%) on MNIST dataset.}\label{Table2:MNIST_cls}
    \parbox{\textwidth}{
    \begin{tabular}{l  c  c }
        \hline
        Method & input & Error ate(\%) \\
        \hline
        PointNet~\cite{Charles2017PointNet} & 256$\times2$ & 0.78 \\
        A-SCN~\cite{xie2018attentional} & 256$\times2$ & 0.60 \\
        KC-Net~\cite{shen2018mining} & 256$\times2$ & 0.70 \\
        PointNet++~\cite{Qi2017PointNet} & 512$\times2$ & 0.51 \\
        Kd-Net~\cite{Klokov2017Escape} & 1024$\times2$ & 0.90 \\
        SpiderCNN~\cite{xu2018spidercnn} & - & - \\
        LocalSpecGCN~\cite{wang2018local} & 1024$\times2$ & \textbf{0.42} \\
        \hline
        AGCN & 512$\times2$ & 0.48 \\
        \hline
    \end{tabular}}
\end{table}

\subsection{Part Segmentation}
\label{sec4.3:part_seg}
\begin{table*}[ht]
\centering
\setlength{\abovecaptionskip}{0pt}
\setlength{\belowcaptionskip}{10pt}
\caption{Accuracy (\%) of part segmentation results on ShapeNet part dataset.}\label{Table3:part_segmentation}
\scalebox{0.68}{
\begin{tabular}{ l c c | c  c  c  c  c  c  c  c  c  c  c  c  c  c  c  c  }
    \cline{1-19}
    & Cat. & Ins. & air & bag & cap & car & chair & ear & guitar & knife & lamp & laptop & motor & mug & pistol & rocket & skate & table\\
    & mIoU & mIoU & plane &  &  &  &  & phone &  &  &  &  & bike &  &  &  & board & \\
	\hline
    \#shapes & & &2690 &76 &55 &898 & 3758 &69 &787 &392 &1547 &451 &202 &184 &283 &66 &152 &5271 \\
    \hline
	PointNet~\cite{Charles2017PointNet} &80.4 &83.7 & 83.4 &78.7 &82.5 &74.9 &89.6 &73.0 &91.5 &85.9 &80.8 &95.3 &65.2 &93.0 &81.2 &57.9 &72.8 &80.6 \\
    PointNet++~\cite{Qi2017PointNet} &81.9 &85.1 &82.4 &79.0 &\textbf{87.7} &77.3 &\textbf{90.8} &71.8 &91.0 &85.9 &83.7 &95.3 &71.6 &94.1 &81.3 &58.7 &76.4 &82.6 \\
    Kd-Net~\cite{Klokov2017Escape} &77.4 &82.3 &80.1 &74.6 &74.3 &70.3 &88.6 &73.5 &90.2 &87.2 &81.0 &94.9 &57.4 &86.7 &78.1 &51.8 &69.9 &80.3 \\
    SpiderCNN~\cite{xu2018spidercnn} &82.4 &85.3 &83.5 &81.0 &87.2 &77.5 &90.7 &\textbf{76.8} &91.1 &87.3 &83.3 &95.8 &70.2 &93.5 &82.7 &59.7 &75.8 &\textbf{82.8} \\
    SynSpecCNN~\cite{yi2017syncspeccnn} &82.0 &84.7 &81.6 &\textbf{81.7} &81.9 &75.2 &90.2 &74.9 &\textbf{93.0} &86.1 &\textbf{84.7} &95.6 &66.7 &92.7 &81.6 &60.6 &\textbf{82.9} &82.1 \\
    A-SCN~\cite{xie2018attentional} &81.8 &84.6 &\textbf{83.8} &80.8 &83.5 &\textbf{79.3} &90.5 & 69.8 &91.7 &86.5 &82.9 &\textbf{96.0} &69.2 &93.8 &82.5 &62.9 &74.4 &80.8 \\
    KC-Net~\cite{shen2018mining} &82.2 &84.7 &82.8 &81.5 &86.4 &77.6 &90.3 &\textbf{76.8} &91.0 &87.2 &84.5 &95.5 &69.2 &\textbf{94.4} &81.6 &60.1 &75.2 &81.3 \\
    \hline
    AGCN &\textbf{82.6} &\textbf{85.4} &83.3 &79.3 &87.5 &78.5 &90.7 &76.5 &91.7 &\textbf{87.8} &\textbf{84.7} &95.7 &\textbf{72.4} &93.2 &\textbf{84.0} &\textbf{63.7} &76.4 &82.5 \\
	\hline
\end{tabular}}
\end{table*}

\begin{figure*}[tb]
    \centering
	\begin{center}
		\includegraphics[width=\linewidth]{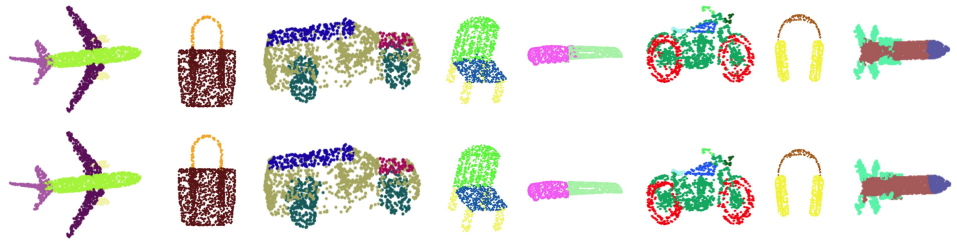}
	\end{center}
	\caption{Part segmentation results. The first row represents the ground truth (GT), and the second row represents our predicted
result.}
	\label{Figure5:part_seg}
\end{figure*}
We evaluated our model for part segmentation on ShapeNet part dataset~\cite{chang2015shapenet}, which contains 16,881 shapes from 16
classes and 50 parts in total. The point of each object is assigned a part label. We use data sets provided by PointNet++
~\cite{Qi2017PointNet} and employ the same experimental setup for training and testing. The task of part segmentation is to predict the
part category of each point, which can be regarded as a point-by-point classification problem.\\
\indent We use the coordinates of the point clouds as well as the surface normal as input to the network. In the stage of local structural
feature learning, we sample $M$=384 nodes to form a set $\mathbf{S}$, and extract $L$=16 nearest points for each node $s_j$ to
build local point set $g_j$. As described in Section~\ref{sec3.4:encoder-decoder}, In the encoder, we stack a 3-layer point attention
layer and build an 8-NN graph for each point attention layer. In addition, we build an 8-NN graph for the point set $\mathbf{S}$ as the input
to the global point graph. Other hyper-parameters are the same as in Section~\ref{sec4.1:3D_cls}.\\
\indent We use intersection-over-union (IoU) to evaluate our our network, the same as PointNet++~\cite{Qi2017PointNet}. The Overall
average instance mIoU(Ins. mIoU) is calculated by averaging IoUs of all the shape instances and the overall average category mIoU(Cat. mIoU) is calculated by averaging over 16 categories. Results are shown in Table~\ref{Table3:part_segmentation},
we can see that with the attention-based encoder-decoder structure, we have achieved good segmentation results for most categories, and
some of the segmentation results are shown in Figure~\ref{Figure5:part_seg}.

\subsection{Semantic Segmentation}
\label{sec4.4:sem_seg}
We evaluate our network on semantic scene segmentation using S3DIS dataset~\cite{Armeni20163D}. S3DIS contains 3D scans from Matterport
scanners in 6 areas including 271 rooms. Each point in the scene point clouds is annotated with one of the semantic labels from 13
categories. We use the same strategy used in PointNet~\cite{Charles2017PointNet} and A-SCN~\cite{xie2018attentional}. The data firstly
split points by room, and then sample rooms into blocks with area 1m by 1m, each block contains 4096 points.\\
\begin{table}[ht]
\setlength{\abovecaptionskip}{0pt}
\setlength{\belowcaptionskip}{10pt}
\caption{6-fold cross validation results on S3DIS dataset.}\label{Table4:sem_seg_acc}
\parbox{.45\textwidth}{
    \begin{tabular}{l  c  c }
        \hline
        Method & mean IoU & Overall \\
        & & accuracy(\%)\\
        \hline
        PointNet~\cite{Charles2017PointNet} &47.71 &78.62 \\
        A-SCN~\cite{xie2018attentional} &52.72 &81.59 \\
        SEGCloud~\cite{tchapmi2017segcloud} &48.92 &- \\
        G+RCU~\cite{engelmann2017exploring} &49.7 &81.1 \\
        RSNet~\cite{huang2018recurrent} &56.47 &- \\
        Engelmann et al.~\cite{engelmann2018know} &\textbf{58.27} &83.95 \\
        \hline
        AGCN &56.63 &\textbf{84.13} \\
        \hline
    \end{tabular}}
\end{table}

\begin{table*}[h]
\centering
\setlength{\abovecaptionskip}{0pt}
\setlength{\belowcaptionskip}{10pt}
\caption{IoU(\%) per class on the S3DIS dataset.}\label{Table5:sem_seg_perclass}
\scalebox{0.8}{
    \begin{tabular}{  l  c  c  c  c  c  c  c  c  c  c  c  c  c }
        \cline{1-14}
        Method &ceiling &floor &wall &beam &column &window &door &chair &table &bookcase &sofa &board &clutter\\
    	\hline
    	SEGCloud~\cite{tchapmi2017segcloud} &90.06 &96.05 &69.86 &0.00 &18.37 &38.35 &23.12 &\textbf{78.59} &\textbf{70.40} &\textbf{58.42} &40.88 &12.96 &41.06\\
        RSNet~\cite{huang2018recurrent} &\textbf{92.48} &92.83 &\textbf{78.56} &32.75 &34.37 &51.62 &\textbf{68.11} &59.72 &60.13 &16.42 &\textbf{50.22} &\textbf{44.85} &\textbf{52.03}\\
        G+RCU~\cite{engelmann2017exploring} &90.3 &92.1 &67.9 &44.7 &24.2 &52.3 &51.2 &58.1 &47.4 &6.9 &39.0 &30.0 &41.9\\
        Engelmann et al.~\cite{engelmann2018know} &92.1 &\textbf{95.0} &72.0 &33.5 &15.0 &46.5 &60.9 &65.1 &69.5 &56.8 &38.2 &6.9 &51.3\\
        \hline
        AGCN &91.37 &94.62 &76.12 &\textbf{54.93} &\textbf{35.23} &\textbf{56.71} &57.69 &62.61 &55.94 &19.37 &46.57 &37.37 &47.64 \\
    	\hline
    \end{tabular}}
\end{table*}

\begin{figure}[!ht]
  \begin{minipage}[c]{0.48\textwidth}
  \includegraphics[width=\textwidth]{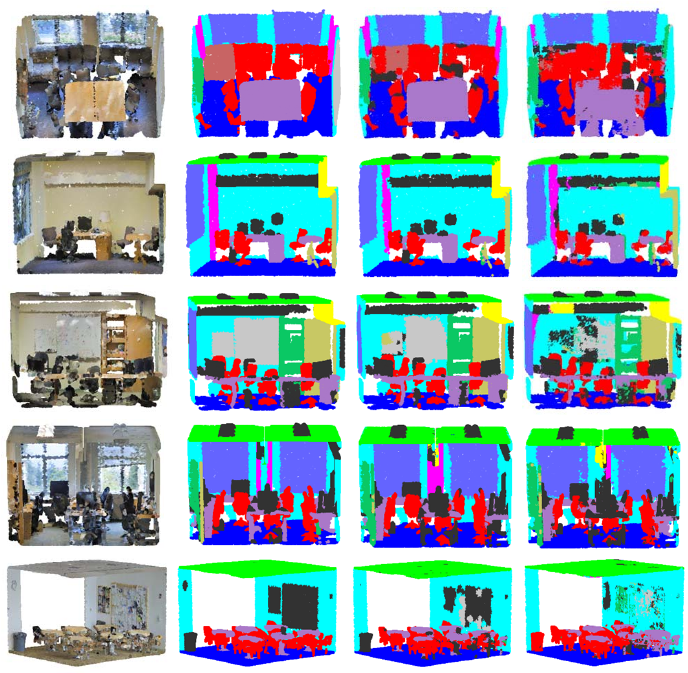}
  \end{minipage}\hfill
  \begin{minipage}[c]{0.48\textwidth}
	\caption{Semantic segmentation results. From left to right: original input scenes, ground truth segmentation, our segmentation
results, PointNet~\cite{Charles2017PointNet} segmentation results.}
	\label{Figure6:sem_seg}
  \end{minipage}
\end{figure}
\indent The input for each point is a 9-dimensional vector (including the xyz, RGB, and the normalized room location). In the stage of
local structural feature learning, we sample $M$=512 nodes to form a set $\mathbf{S}$, and extract $L$=16 nearest points for
each node $s_j$ to build local point set $g_j$. As described in Section~\ref{sec3.4:encoder-decoder}, In the encoder, we stack a 3-layer point attention layer and build an 8-NN graph for each point attention layer. In addition, we build an 8-NN graph for the point set $\mathbf{S}$ as the input to the global point graph.\\
\indent The 6-fold cross validation results of our method are shown in Table~\ref{Table4:sem_seg_acc} and the scores of per class IoU in Table~\ref{Table5:sem_seg_perclass}. Through the encoder-decoder structure, the mean IoU of our model is 56.63\% and the overall accuracy is 84.13\%. Some of the experimental results are shown in Figure~\ref{Figure6:sem_seg}. we can see that, compared to PointNet
~\cite{Charles2017PointNet}, our segmentation results are smoother and the result of segmentation obtained in some flat areas is more
uniform.

\section{Discussion}
\subsection{Influence of different inputs on Network Stability}
\label{sec5.1:stability}
\begin{figure}[!ht]
  \begin{minipage}[c]{0.48\textwidth}
  \includegraphics[width=\textwidth]{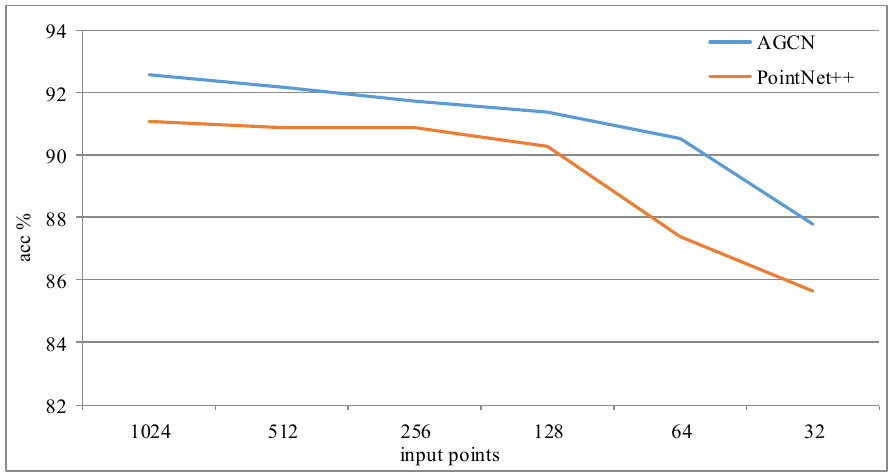}
  \end{minipage}\hfill
  \begin{minipage}[c]{0.48\textwidth}
	\caption{Accuracy with different number of input points on ModelNet40.}
	\label{Figure7:Stability}
  \end{minipage}
\end{figure}

We evaluate the effect of different points on AGCN. Following the settings in Section~\ref{sec4.1:3D_cls}, different numbers of points
and corresponding normals were used as input to train our network and PointNet++~\cite{Qi2017PointNet}. The experimental results are
shown in Figure~\ref{Figure7:Stability}. The accuracy of our network is 87.82\% when it is reduced to 32 points.

\subsection{Effectiveness of global point graph}
\label{sec5.2:w/o_global_point_graph}
\begin{table}[h]
\setlength{\abovecaptionskip}{0pt}
\setlength{\belowcaptionskip}{10pt}
\caption{Accuracy on ModelNet40 with or without global point graph.}\label{Table6:global_point_graph}
\parbox{.5\textwidth}{
	\begin{tabular}{ l  c }
        \cline{1-2}
        With or without global point graph & accuracy(\%)\\
        \hline
        With global point graph &92.61 \\
		\hline
		Without global point graph &90.54 \\
		\hline
	\end{tabular}}
\end{table}
To evaluate the effectiveness of global point graph, we trained two networks (with/without global point graph) on the ModelNet40
classification task. The network settings are the same as the experiments in Section~\ref{sec4.1:3D_cls}. Table
~\ref{Table6:global_point_graph} shows the results of the experiment. It can be seen that the use of global point graph has greatly
improved the network (by nearly 2\%) compared to networks that without global point graph.

\subsection{Visualize point attention layer}
\label{sec5.3:visual_point_attention_layer}
\begin{figure}[!ht]
  \begin{minipage}[c]{0.48\textwidth}
  \includegraphics[width=\textwidth]{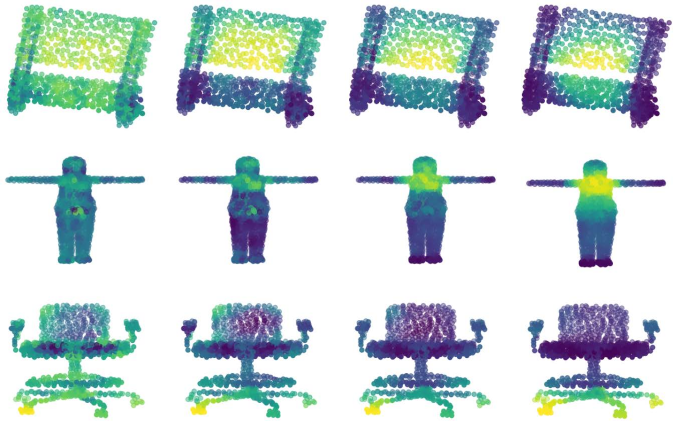}
  \end{minipage}\hfill
  \begin{minipage}[c]{0.48\textwidth}
	\caption{Visualize point attention layer. The first column shows local structural features, from the second column to the
fourth column, we stack 3 layers of point attention layer. We randomly sample one point, and the brighter the color of these points, the
higher the correlation between the features of these points.}
	\label{Figure8:visual_layer}
  \end{minipage}
\end{figure}

We visualize different layers learned from ModelNet40, as illustrated in Figure~\ref{Figure8:visual_layer}. It can be observed that the
features obtained by local structure learning are sparse, but as the depth of the network increases, the distribution of features is close to a cluster, indicating that our point attention layer can effectively aggregate local information. The feature learning from local to global is achieved.

\subsection{Model size and Timing}
\label{sec5.4:size_and_time}
\begin{table}[h]
\setlength{\abovecaptionskip}{0pt}
\setlength{\belowcaptionskip}{10pt}
\caption{Model size and train/inference time. Our networks were tested with an NVIDIA GTX 1080 GPU and an Intel i7-3770K @
3.5 GHz 4 cores CPU. "M" stands for million. The train/inference time is for per batch.}\label{Table7:size_time}
\parbox{.48\textwidth}{
	\begin{tabular}{ l  c  c }
        \cline{1-3}
        Method &Parameters &Train/Inference time \\
        & &(per batch) \\
		\hline
		PointNet~\cite{Charles2017PointNet} &3.48M &0.083/0.024 s \\
        PointNet++~\cite{Qi2017PointNet} &1.48M &0.526/0.203 s \\
        SpiderCNN~\cite{xu2018spidercnn} &3.23M &0.332/0.145 s \\
		\hline
        AGCN &2.03M &0.076/0.033 s \\
        \hline
	\end{tabular}}
\end{table}
We recorded the number of parameters of different networks and the time of training/inference on the tasks of the ModelNet40 classification. For fair comparison, we set the batch size to 16, 1024 points as input to the network, other settings for experiments are the same as in Section~\ref{sec4.1:3D_cls}. We statistic network parameters, and average training/inference time for each batch. The experimental results available in Table~\ref{Table7:size_time}, we can see that our network can exceed or approach the latest methods. Although our parameters are more than PointNet++~\cite{Qi2017PointNet}, thanks to our point attention layer, which only focuses on neighboring points, the amount of computation of the network is reduced.

\subsection{Visualize the local patterns}
\label{sec5.5:local_pattern}
\begin{figure}[!ht]
  \begin{minipage}[c]{0.48\textwidth}
  \includegraphics[width=\textwidth]{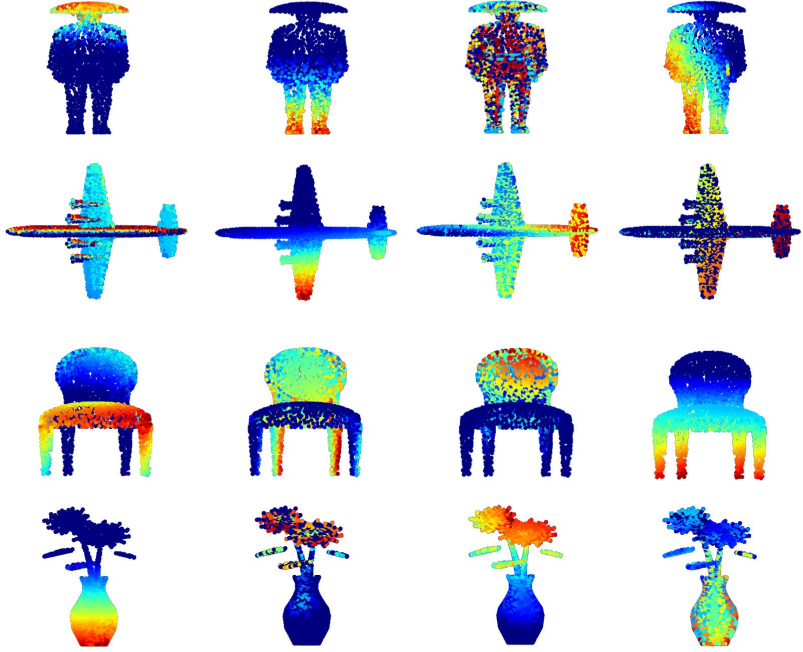}
  \end{minipage}\hfill
  \begin{minipage}[c]{0.48\textwidth}
	\caption{Visualize the patterns learned from the first layer. Each row represents the same sample, and for each sample we select 4 learned kernels, red color indicates the highest response to the activation of the kernel, and blue color indicates the lowest response to the activation of the kernel.}
	\label{Figure9:local_pattern}
  \end{minipage}
\end{figure}

We visualize the local patterns learned by the kernels of the first layer of our network on ModelNet40. For each point set, we use all
points as input to activate the specific neurons. In Figure~\ref{Figure9:local_pattern}, We can see that the kernel of our network can
learn local patterns very well, such as lines, planes etc.

\section{Conclusion}
In this paper, we present AGCN. We introduced the attention mechanism in the graph structure network to build point attention layer, and learn the relationship between local features. Also, we extended the point attention layer to build encoder-decoder attention network for segmentation tasks. To compensate for the relative information of individual points in the graph network, we introduce an additional global graph structure network. Through some extensive experiments, we can see that point attention layer can effectively model the relationship between local features and achieve local-to-global feature learning. In the future work, we will further improve the point attention layer for 3D semantic analysis.

\section*{Acknowledgement}
This work is partly supported by the National Natural Science Foundation of China (No.61876158) and Sichuan science and technology program (No.19ZDYF2070).

\section*{Reference}
\bibliographystyle{plain}
\bibliography{bibfile}

\begin{thebibliography}{10}

\bibitem{armeni2017joint}
Iro Armeni, Sasha Sax, Amir~R Zamir, and Silvio Savarese.
\newblock Joint 2d-3d-semantic data for indoor scene understanding.
\newblock {\em arXiv preprint arXiv:1702.01105}, 2017.

\bibitem{Armeni20163D}
Iro Armeni, Ozan Sener, Amir~R. Zamir, Helen Jiang, Ioannis Brilakis, Martin
  Fischer, and Silvio Savarese.
\newblock 3d semantic parsing of large-scale indoor spaces.
\newblock In {\em Computer Vision \& Pattern Recognition}, 2016.

\bibitem{chang2015shapenet}
Angel~X Chang, Thomas Funkhouser, Leonidas Guibas, Pat Hanrahan, Qixing Huang,
  Zimo Li, Silvio Savarese, Manolis Savva, Shuran Song, Hao Su, et~al.
\newblock Shapenet: An information-rich 3d model repository.
\newblock {\em arXiv preprint arXiv:1512.03012}, 2015.

\bibitem{Charles2017PointNet}
R.~Qi Charles, Su~Hao, Kaichun Mo, and Leonidas~J. Guibas.
\newblock Pointnet: Deep learning on point sets for 3d classification and
  segmentation.
\newblock In {\em IEEE Conference on Computer Vision \& Pattern Recognition},
  2017.

\bibitem{cheng2016long}
Jianpeng Cheng, Li~Dong, and Mirella Lapata.
\newblock Long short-term memory-networks for machine reading.
\newblock {\em arXiv preprint arXiv:1601.06733}, 2016.

\bibitem{Dai2017ScanNet}
Angela Dai, Angel~X. Chang, Manolis Savva, Maciej Halber, Thomas Funkhouser,
  and Matthias Nie?ner.
\newblock Scannet: Richly-annotated 3d reconstructions of indoor scenes.
\newblock 2017.

\bibitem{Defferrard2016Convolutional}
Micha?l Defferrard, Xavier Bresson, and Pierre Vandergheynst.
\newblock Convolutional neural networks on graphs with fast localized spectral
  filtering.
\newblock 2016.

\bibitem{engelmann2017exploring}
Francis Engelmann, Theodora Kontogianni, Alexander Hermans, and Bastian Leibe.
\newblock Exploring spatial context for 3d semantic segmentation of point
  clouds.
\newblock In {\em Proceedings of the IEEE International Conference on Computer
  Vision}, pages 716--724, 2017.

\bibitem{engelmann2018know}
Francis Engelmann, Theodora Kontogianni, Jonas Schult, and Bastian Leibe.
\newblock Know what your neighbors do: 3d semantic segmentation of point
  clouds.
\newblock In {\em European Conference on Computer Vision}, pages 395--409.
  Springer, 2018.

\bibitem{Hackel2017Semantic3D}
Timo Hackel, Nikolay Savinov, Lubor Ladicky, Jan~D. Wegner, Konrad Schindler,
  and Marc Pollefeys.
\newblock Semantic3d.net: A new large-scale point cloud classification
  benchmark.
\newblock 2017.

\bibitem{he2016deep}
Kaiming He, Xiangyu Zhang, Shaoqing Ren, and Jian Sun.
\newblock Deep residual learning for image recognition.
\newblock In {\em Proceedings of the IEEE conference on computer vision and
  pattern recognition}, pages 770--778, 2016.

\bibitem{huang2018recurrent}
Qiangui Huang, Weiyue Wang, and Ulrich Neumann.
\newblock Recurrent slice networks for 3d segmentation of point clouds.
\newblock In {\em Proceedings of the IEEE Conference on Computer Vision and
  Pattern Recognition}, pages 2626--2635, 2018.

\bibitem{ioffe2015batch}
Sergey Ioffe and Christian Szegedy.
\newblock Batch normalization: Accelerating deep network training by reducing
  internal covariate shift.
\newblock {\em arXiv preprint arXiv:1502.03167}, 2015.

\bibitem{Kipf2016Semi}
Thomas~N. Kipf and Max Welling.
\newblock Semi-supervised classification with graph convolutional networks.
\newblock 2016.

\bibitem{Klokov2017Escape}
Roman Klokov and Victor Lempitsky.
\newblock Escape from cells: Deep kd-networks for the recognition of 3d point
  cloud models.
\newblock In {\em 2017 IEEE International Conference on Computer Vision
  (ICCV)}, 2017.

\bibitem{li2016fpnn}
Yangyan Li, Soeren Pirk, Hao Su, Charles~R Qi, and Leonidas~J Guibas.
\newblock Fpnn: Field probing neural networks for 3d data.
\newblock In {\em Advances in Neural Information Processing Systems}, pages
  307--315, 2016.

\bibitem{long2015fully}
Jonathan Long, Evan Shelhamer, and Trevor Darrell.
\newblock Fully convolutional networks for semantic segmentation.
\newblock In {\em Proceedings of the IEEE conference on computer vision and
  pattern recognition}, pages 3431--3440, 2015.

\bibitem{Maturana2015VoxNet}
Daniel Maturana and Sebastian Scherer.
\newblock Voxnet: A 3d convolutional neural network for real-time object
  recognition.
\newblock In {\em IEEE/RSJ International Conference on Intelligent Robots \&
  Systems}, 2015.

\bibitem{Noh2015Learning}
Hyeonwoo Noh, Seunghoon Hong, and Bohyung Han.
\newblock Learning deconvolution network for semantic segmentation.
\newblock In {\em IEEE International Conference on Computer Vision}, 2015.

\bibitem{Novotny2017Learning}
David Novotny, Diane Larlus, and Andrea Vedaldi.
\newblock Learning 3d object categories by looking around them.
\newblock 2017.

\bibitem{Qi2016Volumetric}
Charles~R. Qi, Su~Hao, Matthias Niessner, Angela Dai, and Leonidas~J. Guibas.
\newblock Volumetric and multi-view cnns for object classification on 3d data.
\newblock 2016.

\bibitem{Qi2017PointNet}
Charles~R. Qi, Yi~Li, Su~Hao, and Leonidas~J. Guibas.
\newblock Pointnet++: Deep hierarchical feature learning on point sets in a
  metric space.
\newblock 2017.

\bibitem{Riegler2017OctNet}
Gernot Riegler, Ali~Osman Ulusoy, and Andreas Geiger.
\newblock Octnet: Learning deep 3d representations at high resolutions.
\newblock 2017.

\bibitem{shen2018mining}
Yiru Shen, Chen Feng, Yaoqing Yang, and Dong Tian.
\newblock Mining point cloud local structures by kernel correlation and graph
  pooling.
\newblock In {\em Proceedings of the IEEE conference on computer vision and
  pattern recognition}, pages 4548--4557, 2018.

\bibitem{Simonovsky2017Dynamic}
Martin Simonovsky and Nikos Komodakis.
\newblock Dynamic edge-conditioned filters in convolutional neural networks on
  graphs.
\newblock 2017.

\bibitem{su2015multi}
Hang Su, Subhransu Maji, Evangelos Kalogerakis, and Erik Learned-Miller.
\newblock Multi-view convolutional neural networks for 3d shape recognition.
\newblock In {\em Proceedings of the IEEE international conference on computer
  vision}, pages 945--953, 2015.

\bibitem{tchapmi2017segcloud}
Lyne Tchapmi, Christopher Choy, Iro Armeni, JunYoung Gwak, and Silvio Savarese.
\newblock Segcloud: Semantic segmentation of 3d point clouds.
\newblock In {\em 2017 International Conference on 3D Vision (3DV)}, pages
  537--547. IEEE, 2017.

\bibitem{vaswani2017attention}
Ashish Vaswani, Noam Shazeer, Niki Parmar, Jakob Uszkoreit, Llion Jones,
  Aidan~N Gomez, {\L}ukasz Kaiser, and Illia Polosukhin.
\newblock Attention is all you need.
\newblock In {\em Advances in Neural Information Processing Systems}, pages
  5998--6008, 2017.

\bibitem{wang2018local}
Chu Wang, Babak Samari, and Kaleem Siddiqi.
\newblock Local spectral graph convolution for point set feature learning.
\newblock In {\em Proceedings of the European Conference on Computer Vision
  (ECCV)}, pages 52--66, 2018.

\bibitem{wang2015voting}
Dominic~Zeng Wang and Ingmar Posner.
\newblock Voting for voting in online point cloud object detection.
\newblock In {\em Robotics: Science and Systems}, volume~1, pages 10--15607,
  2015.

\bibitem{wang2018dynamic}
Yue Wang, Yongbin Sun, Ziwei Liu, Sanjay~E Sarma, Michael~M Bronstein, and
  Justin~M Solomon.
\newblock Dynamic graph cnn for learning on point clouds.
\newblock {\em arXiv preprint arXiv:1801.07829}, 2018.

\bibitem{Wu20143D}
Zhirong Wu, Shuran Song, Aditya Khosla, Fisher Yu, Linguang Zhang, Xiaoou Tang,
  and Jianxiong Xiao.
\newblock 3d shapenets: A deep representation for volumetric shapes.
\newblock 2014.

\bibitem{xie2018attentional}
Saining Xie, Sainan Liu, Zeyu Chen, and Zhuowen Tu.
\newblock Attentional shapecontextnet for point cloud recognition.
\newblock In {\em Proceedings of the IEEE Conference on Computer Vision and
  Pattern Recognition}, pages 4606--4615, 2018.

\bibitem{xu2018spidercnn}
Yifan Xu, Tianqi Fan, Mingye Xu, Long Zeng, and Yu~Qiao.
\newblock Spidercnn: Deep learning on point sets with parameterized
  convolutional filters.
\newblock In {\em Proceedings of the European Conference on Computer Vision
  (ECCV)}, pages 87--102, 2018.

\bibitem{yi2017syncspeccnn}
Li~Yi, Hao Su, Xingwen Guo, and Leonidas~J Guibas.
\newblock Syncspeccnn: Synchronized spectral cnn for 3d shape segmentation.
\newblock In {\em Proceedings of the IEEE Conference on Computer Vision and
  Pattern Recognition}, pages 2282--2290, 2017.

\end{thebibliography}

\end{document}